\documentclass[letterpaper, 10 pt, journal, twoside]{IEEEtran}
\usepackage{graphicx, tipa}
\usepackage{amsmath,amssymb,amsfonts}
\usepackage[caption=false]{subfig}
\usepackage[ruled,linesnumbered, noend]{algorithm2e}
\usepackage{dirtytalk}
\usepackage{hyperref}
\usepackage{gensymb}
\usepackage{tabularx}
\usepackage{multirow}
\usepackage[short]{optidef}
\usepackage{booktabs}
\usepackage{siunitx}
\usepackage{titlesec}
\newcommand{\arc}[1]{{%
  \setbox9=\hbox{#1}%
  \ooalign{\resizebox{\wd9}{\height}{\texttoptiebar{\phantom{A}}}\cr#1}}}
\makeatletter
\newcommand{\removelatexerror}{\let\@latex@error\@gobble}
\newcommand\Tstrut{\rule{0pt}{2.0ex}}         

\makeatother
\SetKwComment{Comment}{$\triangleright$\ }{}

\IEEEoverridecommandlockouts
\title{\huge 
A vision-based autonomous UAV inspection framework for unknown tunnel construction sites with dynamic obstacles\\
        
\thanks{Manuscript received: January 19, 2023; Revised:
April 27, 2023; Accepted: June 1, 2023. This paper was recommended for publication by Editor Pauline Pounds upon evaluation of the Associate Editor and Reviewers’ comments.

Zhefan Xu, Baihan Chen, Xiaoyang Zhan, Yumeng Xiu, Christopher Suzuki, and Kenji Shimada are with the Department of Mechanical Engineering, Carnegie Mellon University, 5000 Forbes Ave, Pittsburgh, PA, 15213, USA.,
         {\tt\small zhefanx@andrew.cmu.edu}
         
Digital Object Identifier (DOI): see top of this page.}}

\author{\IEEEauthorblockN{Zhefan Xu, Baihan Chen, Xiaoyang Zhan, Yumeng Xiu, Christopher Suzuki, and Kenji Shimada}}

\begin{document}

\markboth{IEEE Robotics and Automation Letters. Preprint Version. Accepted June, 2023}
{Xu \MakeLowercase{\textit{et al.}}: A vision-based autonomous UAV inspection framework for
unknown tunnel construction sites with dynamic obstacles} 

\maketitle

\noindent \begin{abstract}
Tunnel construction using the drill-and-blast method requires the 3D measurement of the excavation front to evaluate underbreak locations. Considering the inspection and measurement task's safety, cost, and efficiency, deploying lightweight autonomous robots, such as unmanned aerial vehicles (UAV), becomes more necessary and popular. Most of the previous works use a prior map for inspection viewpoint determination and do not consider dynamic obstacles. To maximally increase the level of autonomy, this paper proposes a vision-based UAV inspection framework for dynamic tunnel environments without using a prior map. Our approach utilizes a hierarchical planning scheme, decomposing the inspection problem into different levels. The high-level decision maker first determines the task for the robot and generates the target point. Then, the mid-level path planner finds the waypoint path and optimizes the collision-free static trajectory. Finally, the static trajectory will be fed into the low-level local planner to avoid dynamic obstacles and navigate to the target point. Besides, our framework contains a novel dynamic map module that can simultaneously track dynamic obstacles and represent static obstacles based on an RGB-D camera. After inspection, the Structure-from-Motion (SfM) pipeline is applied to generate the 3D shape of the target. To our best knowledge, this is the first time autonomous inspection has been realized in unknown and dynamic tunnel environments. Our flight experiments in a real tunnel prove that our method can autonomously inspect the tunnel excavation front surface. The video is available at: \url{https://youtu.be/MSNp-hg9RCQ}. Our software is available on GitHub\footnote{\url{https://github.com/Zhefan-Xu/CERLAB-UAV-Autonomy}} as an open-source ROS package.
\end{abstract}

\begin{IEEEkeywords}
Field Robotics, Motion and Path Planning, Perception and Autonomy, Robotics and Automation in Construction
\end{IEEEkeywords}

\section{Introduction}
Drilling and blasting is a common tunnel construction and excavation method. The main cycle of this method includes steps such as drilling for explosives, blasting, measuring underbreaks, and spraying concrete. Among these steps, measuring underbreaks in the tunnel excavation front is dangerous for workers because of the potential falling rocks. With the emergence of lightweight unmanned aerial vehicles, the robot becomes suitable for handling measurement and inspection tasks as it can avoid potential human dangers and inspect unreachable locations. Consequently, an autonomous inspection framework is essential to improve the safety and efficiency of underbreaks measurement and tunnel construction. 
\begin{figure}[t] 
    \centering
    \includegraphics[scale=0.815]{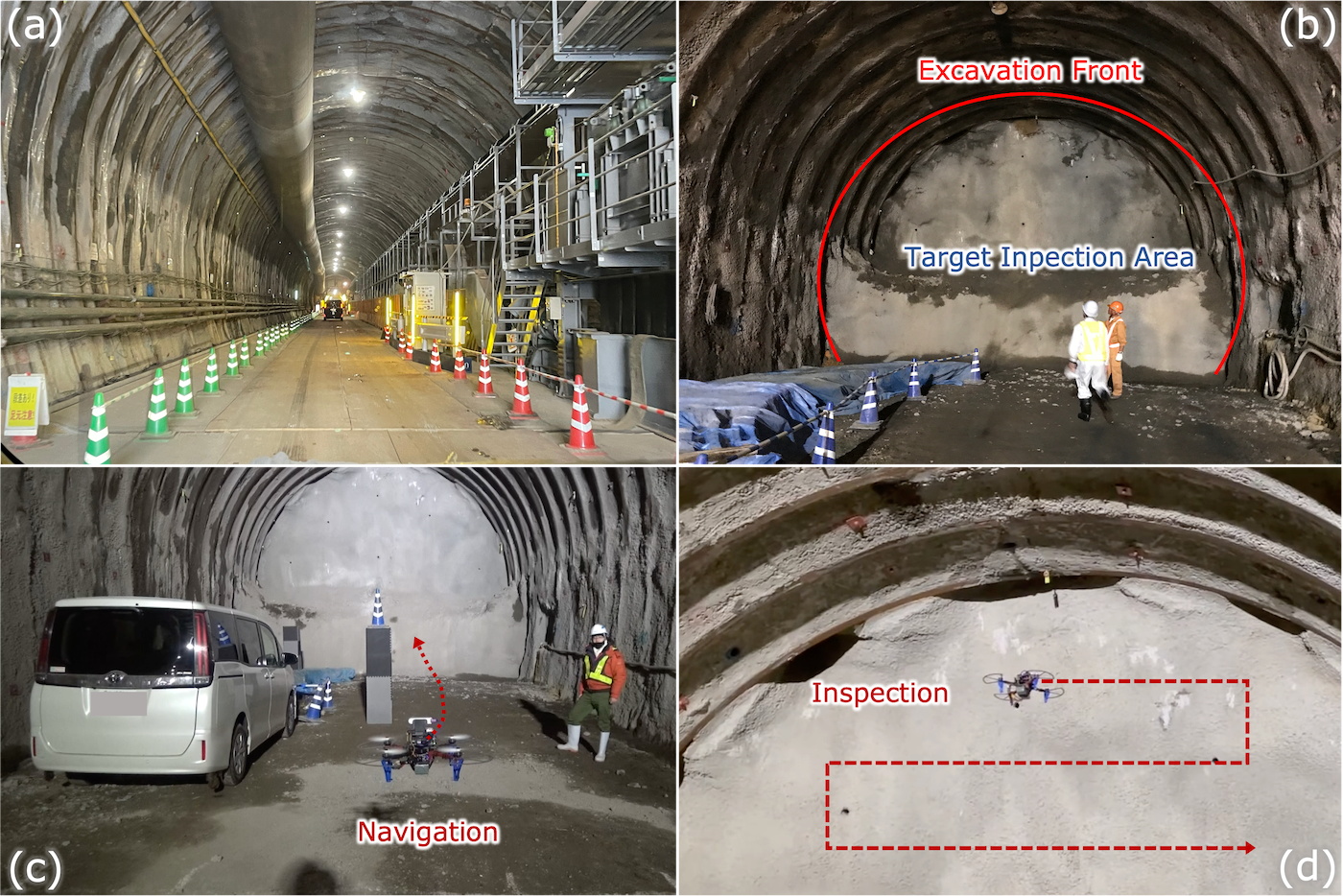}
    \caption{Illustration of UAV navigating and inspecting the excavation front in the tunnel environment. (a) The tunnel under construction. (b) The target inspection area (the excavation front). (c) The robot navigates toward the inspection target and avoids obstacles. (d) The robot inspects the target area.}
    \label{intro_figure}
\end{figure}

There are two main challenges of autonomous UAV inspection in tunnel environments. First, since the tunnel environments under construction are changing with time, it is unlikely to
have update-to-date maps of huge construction vehicles and equipment nearby the excavation front. In this way, the robot should be able to navigate from arbitrary positions in the tunnel towards the excavation front area (i.e., the end of the tunnel) based on the onboard sensing. Previous works of the sampling-based unknown exploration \cite{RH-NBV}\cite{AEP}\cite{TSDF}\cite{DEP} can make the robot successfully navigate and map unknown environments with the onboard sensor and applies this exploration method to the unknown tunnel inspection \cite{tunnel_inspection}. However, because their approaches only utilize the explored map information to randomly sample viewpoints, the output trajectory could be zigzag and over-conservative, making navigation less efficient. The second challenge comes from the moving workers and machines in tunnels, as the robot should track them and avoid them safely. Even though some recent research \cite{reactiveUV}\cite{visionCCMPC}\cite{ViGO} has investigated the UAV dynamic obstacle avoidance problems, their local planning strategies without global path fusion make them insufficient for complex inspection tasks in tunnel environments, which contain complicated static structures and unpredictable dynamic obstacles. 

To solve these issues, this paper proposes a vision-based autonomous UAV inspection framework for unknown and dynamic tunnel environments. We develop a small, lightweight quadcopter with an RGB-D camera for safely sharing and operating with vehicles, equipment and workers in the tunnel. The proposed approach utilizes the hierarchical planning method decomposing the entire inspection planning into high, mid, and low levels. The current task is determined at the high planning level to generate the goal position for navigation and exploration. Then, the mid-level planner will find and optimize a smooth trajectory toward the goal based on the static obstacle information from the incrementally built map. Finally, at the low level, our vision-aided gradient-based planner is applied to locally optimize the trajectory for avoiding dynamic obstacles. The example tunnel environment and the autonomous robot using the proposed method are shown in Fig. \ref{intro_figure}.
The main contributions and novelties of this work are:
\begin{itemize}
  \item \textbf{Autonomous UAV system for inspecting unknown and dynamic tunnels:} We design a novel autonomous UAV system tailored to tunnel inspection. The system adopts our hierarchical planning framework and the dynamic-obstacle-aware map representation to achieve safe inspection in unknown and dynamic tunnel environments. 
  \item \textbf{Lightweight U-depth dynamic obstacle detector:} The paper introduces our lightweight U-depth map-based detector for dynamic obstacle detection. The detected bounding boxes of dynamic obstacles can be merged with the traditional occupancy map to form the dynamic map for safe static and dynamic obstacle avoidance.
  \item \textbf{Gradient-based dynamic obstacle avoidance:} We propose the circle-based guide-point method and the receding horizon distance field to efficiently compute the trajectory optimization gradients for obstacle avoidance. 
    \item \textbf{Large-scale tunnel inspection experiments:} The entire system was verified using our customized quadcopter. The experiment was conducted in a tunnel under construction with a cross-section size of 8 meters in height and 8 meters in width in Japan. The experiment demonstrates that our system can safely inspect the tunnel and output 3D reconstruction results for tunnel construction.
\end{itemize}

\section{Related Work}
This section first discusses the recent trends and approaches in construction site inspection by autonomous UAVs. Then, relevant works on the key challenges of tunnel inspection (i.e., exploration and dynamic obstacle avoidance) are reviewed. 

There are mainly two categories of construction site and building inspection methods: model-based and non-model-based methods. For the model-based methods, the inspection target model is usually available, and the planner generates a set of optimal viewpoints based on the provided model. In \cite{bridge_GTSP}, the target bridge is first partitioned into surfaces with inspection nodes, and their GTSP solver is then applied to find optimal paths for inspection. Similarly, some works use the BIM model to find viewpoints of interest (VPI) and solve the path-planning problem using the TSP-based method \cite{model_based_1}\cite{model_based_2}. However, the target model can be unavailable for tunnel inspection, so the robot can only rely on the onboard sensors. In this way, the reactive methods are proposed for unknown tunnel navigation using the lidar points measurement  \cite{lidar_tunnel1}\cite{lidar_tunnel2}. Their methods can navigate tunnels of arbitrary shapes but do not consider obstacle avoidance. Bendris et al. \cite{tunnel_inspection} utilize the sampling-based method to generate viewpoints for unknown exploration and inspection.  Their method can successfully avoid static obstacles but might not be safe for dynamic obstacles due to the long replanning time. Besides, their random sampling strategy in the explored area can lead to zigzag and over-conservative paths for navigation. In \cite{tunnel_reconstruction}, it proposes a 3D reconstruction method for UAV tunnel inspection without the path-planning strategy.  

The unknown exploration problem can be viewed as determining a series of informative viewpoints \cite{nbv_idea}. Yamauchi \cite{first_frontier} first uses the frontier exploration approach, allowing robots to visit the map boundary to gain environment information. Later in \cite{rapid_frontier}, it extends the frontier exploration to high-speed UAVs. Some approach \cite{information_gain} applies the information-theoretic method to evaluate the information gains of viewpoints. Considering the limited computation power of lightweight UAVs, the sampling-based methods \cite{RH-NBV}\cite{AEP}\cite{TSDF}\cite{DEP} have been preferred in recent years. In \cite{RH-NBV}, their RH-NBV planner grows an RRT with the information gains stored in each node. The robot will then follow the highest gain branch in a receding horizon manner. Selin et al. \cite{AEP} combine the RH-NBV with frontier exploration, further improving the exploration efficiency. To save and reuse the computation in each planning iteration, Schmid et al. \cite{TSDF} adopt the RRT* algorithm with the rewiring to incrementally build the tree. With a similar incremental sampling idea in \cite{DEP}, it proposes a PRM-based method for exploration and obstacle avoidance in dynamic environments.

Dynamic obstacle avoidance problem still remains open in recent years. In the reactive-based methods, the robots directly generate control velocities to avoid obstacles. Khatib \cite{apf} constructs the artificial potential field to find the velocity for obstacle avoidance and navigation, and Berg et al. \cite{orca} use linear programming to optimize velocities based on Velocity Obstacle \cite{VO}. These methods require less computation than the trajectory-based methods but might lead to more myopic performance. The trajectory-based methods are more prevalent in UAV planning in recent years. Some \cite{11_chance}\cite{19_chance}\cite{20_chance}\cite{dpmpc} use the model predictive control scheme to generate collision-free trajectories based on the kinematic constraints. In \cite{ViGO}, it utilizes the B-spline optimization to generate collision-free trajectory with vision aided, and Chen et al. \cite{chen2022risk} evaluate trajectory risks using their dual-structure particle map.

\section{Problem Description}
In an unknown tunnel space, $\mathcal{V}_{t} \in \mathbb{R}^3$, with a straight tunnel centerline $\mathcal{C}$ of a finite length, there exists an excavation front (i.e., the target wall for inspection) at the end of the tunnel. Inside the tunnel space  $\mathcal{V}_{t}$, there are different sizes of static obstacles $\mathcal{O}_{\text{static}}$ and dynamic obstacles $\mathcal{O}_{\text{dynamic}}$. A UAV with an onboard depth camera is deployed for the inspection task. Without a prior map $\mathcal{M}$, the robot needs to first navigate toward the excavation front area from an arbitrary position in the space $\mathcal{V}_{t}$, then generate an inspection path to collect RGB images of the inspection target, and finally return to the start location. During the forward navigation and returning period, the robot should avoid all static obstacles $\mathcal{O}_{\text{static}}^{\text{all}}$ and dynamic obstacles $\mathcal{O}_{\text{dynamic}}^{\text{sensor}}$ in its sensor range. The final output of the entire system should be the 3D shape of the inspection target reconstructed using the collected RGB images.

\section{Proposed Method} \label{method}
\begin{figure*}[t] 
    \centering
    \includegraphics[scale=0.505]{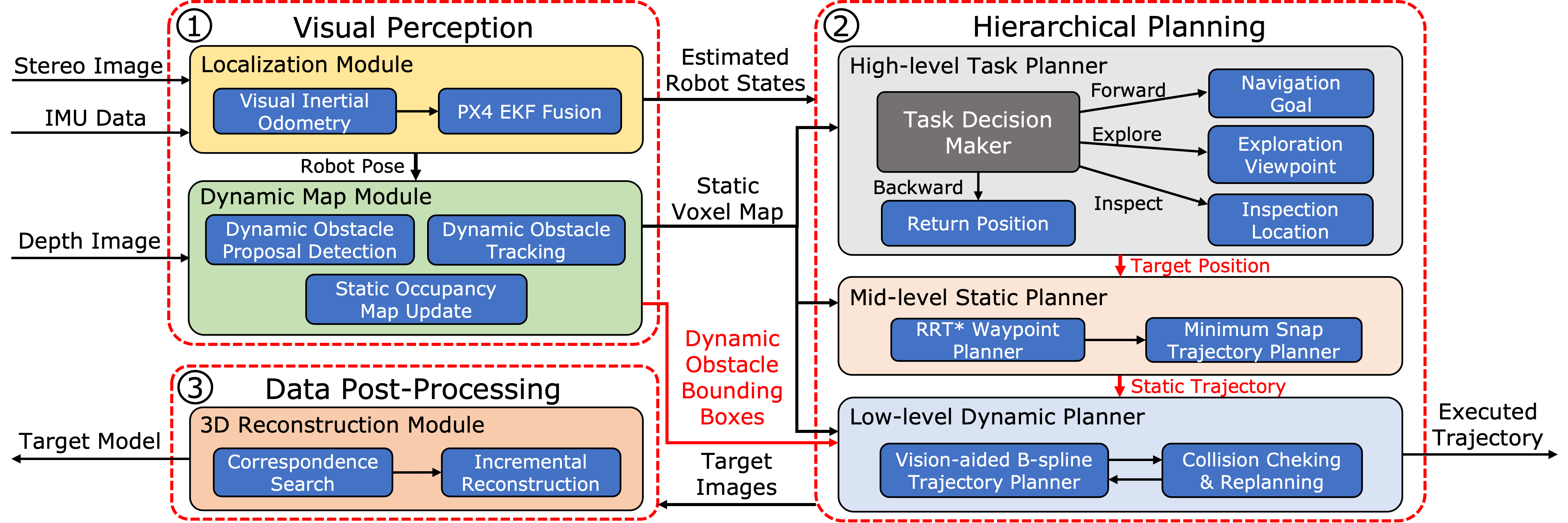}
    \caption{System framework for autonomous inspection. Our proposed framework contains three parts: visual perception, hierarchical planning, and data post-processing. In the visual perception step, the localization module applies the visual-inertial odometry with EKF fusion for state estimation. The dynamic map module builds the static voxel map and tracks dynamic obstacles based on depth images. In the hierarchical planning section, the high-level and mid-level planners use the static voxel map to generate the static trajectory. Then, the low-level planner uses the dynamic obstacle information to optimize the output trajectory for execution. The final data post-processing step takes the images collected from the inspection stage to reconstruct the target model for analysis. }
    \label{system_framework}
\end{figure*}
The proposed inspection framework has three main components shown in Fig. \ref{system_framework}: visual perception, hierarchical planning, and data post-processing. The visual perception step processes the sensor measurements from the onboard depth camera and the inertial measurement unit (IMU). The localization module runs the visual-inertial odometry (VIO) algorithm with the EKF fusion to get robot state estimation. Besides, the dynamic map module utilizes depth images to track dynamic obstacles and update the occupancy information for static obstacles using the voxel map, which will be further discussed in Sec.\ref{perception mapping}. After the perception step, the hierarchical planning section generates collision-free trajectories for the robot to achieve the entire inspection task. Sec. \ref{hierachical planning} will introduce the logic of our hierarchical planning for the tunnel inspection and the task decision maker in the high-level planner. Then, the obstacle avoidance based on the mid-level trajectory planner and low-level dynamic planner will be covered in Sec. \ref{obstacle avoidance}. After finishing the inspection task, the data post-processing step,  mentioned in Sec. \ref{3d reconstruction}, takes the collected target images to perform 3D reconstruction to obtain the target model.

\subsection{Hierarchical Planning and High-level Task Planner} \label{hierachical planning}
Since our inspection problem consists of multiple complicated procedures, applying only one planner cannot efficiently accomplish the entire task. There are mainly three stages of the inspection: (a) approaching the inspection target (i.e., the end of the tunnel), (b) collecting target images, and (c) returning to the start location. Based on the inspection stages, we decompose the problem into the following abstract tasks:
\begin{equation}
\mathcal{S}_{\text{T}} = \{\text{Forward}, \text{Explore}, \text{Inspect}, \text{Return}\},
\end{equation}
where the Forward task aims at approaching the inspection target, the Explore task helps the robot gain local map information for navigation, the Inspect task mode generates the path for collecting target images, and the Return task mode navigates the robot back to the starting location. During the inspection process, the robot constantly alternates the task mode using the proposed task planning algorithm (Alg. \ref{task planning}). For each abstract task, the task planner generates the corresponding goal positions and passes them to the lower-level planners for path planning and trajectory optimization. 

In the beginning stage of task planning (Alg. \ref{task planning}), the task planner sets the robot to the Forward task mode as the robot needs first to approach the tunnel end (Line \ref{initialization}). The task planner runs at a certain replanning frequency to select the current task mode for the robot. Before the robot arrives at the inspection location, the \textbf{Forward Mode} (Lines \ref{start forward mode}-\ref{end forward mode}) lets the robot generate a forward goal with a distance $l$ from the current robot position for navigation. Since, at this stage, the robot does not have a complete environment map and can only rely on the partially built from its flight, it will first try using the partial map to perform local obstacle avoidance to achieve the forward goal (Line \ref{forward local}). Suppose the lower-level planner fails to find a collision-free trajectory due to the lack of environmental knowledge. In that case, the task planner will switch the current task to the \textbf{Explore Mode} to increase the local map information (Lines \ref{start forward to explore}-\ref{end forward mode}). In the Explore Mode, the planner first samples to get the best viewpoints with the highest sensor information gain in the current map then uses the lower-level planner to generate a feasible trajectory for exploration, and finally switches back to the previous task mode (Lines \ref{start explore mode}-\ref{end explore mode}). For the information gain evaluation, refers to \cite{RH-NBV}\cite{AEP}\cite{TSDF}\cite{DEP} for further details. At the start of each replanning iteration, the algorithm checks whether the robot has reached the inspection target (Lines \ref{start check inspection}-\ref{end check inspection}). If the robot detects the inspection target wall, the planner will enter the \textbf{Inspect Mode} and generate a zigzag path for collecting target images. However, when the built map around the target is not detailed enough for the inspection path generation, the planner will switch to the Explore Mode again to increase the explored map range (Lines \ref{start inspect mode}-\ref{end inspect mode}). After finishing collecting images, the planner will enter the \textbf{Return Mode} and navigate back to the start position (Lines \ref{start return mode}-\ref{end inspection}). Note that in the returning step, the robot has already had a sufficient informative map for static obstacles, incrementally built from the forward and explore step, to generate a global trajectory to the origin directly.  

\begin{algorithm}[t] \label{task planning}
\caption{High-level Task Planning Algorithm} 
\SetAlgoNoLine%

$\mathcal{T}_{\text{curr}} \gets \text{Forward Mode}$ \Comment*[r]{initial task} \label{initialization}
$\mathcal{C}_{t} \gets false$ \Comment*[r]{termination condition}
\While{\normalfont{\textbf{not}} $\mathcal{C}_{t}$}{ 
    $\mathcal{C}_{t} \gets \normalfont{\textbf{isInspectionComplete}}()$\; 

    $\mathcal{I}_{\text{reach}} \gets \normalfont{\textbf{reachInspectionTarget}}()$\; \label{start check inspection}
    \If{$\mathcal{I}_{\normalfont{\text{reach}}}$}{
        $\mathcal{T}_{\text{\normalfont{curr}}} \gets \normalfont{\text{Inspect Mode}}$\;\label{end check inspection}
    }
    
    \Switch{$\mathcal{T}_{\text{\normalfont{curr}}}$}{
            \Case{\normalfont{Forward Mode}}{ \label{start forward mode}
                $\mathcal{P}_{\text{\normalfont{goal}}} \gets$ \normalfont{\textbf{getForwardGoal}}()\;
                $\sigma_{\normalfont{\text{traj}}}, \normalfont{\text{success}} \gets$ \textbf{lowerLevelPlanner}($\mathcal{P}_{\text{\normalfont{goal}}}$)\; \label{forward local}
                \If{\normalfont{\textbf{not}} \normalfont{\text{success}}}{ \label{start forward to explore}
                    $\mathcal{T}_{\text{\normalfont{curr}}} \gets \normalfont{\text{Explore Mode}}$\; \label{end forward mode}
                } 
            }
            \Case{\normalfont{Explore Mode}}{ \label{start explore mode}
                $\mathcal{P}_{\text{\normalfont{goal}}} \gets$ \normalfont{\textbf{getBestViewpoint}}()\;
                $\sigma_{\normalfont{\text{traj}}} \gets$ \textbf{lowerLevelPlanner}($\mathcal{P}_{\text{\normalfont{goal}}}$)\; 
                $\mathcal{T}_{\text{\normalfont{curr}}} \gets $ \normalfont{\textbf{getPreviousTaskMode}}()\; \label{end explore mode}
            }
            \Case{\normalfont{Inspect Mode}}{\label{start inspect mode}
                $\sigma_{\normalfont{\text{traj}}}, \normalfont{\text{success}} \gets$ \textbf{getInspectionPath}()\;
                \eIf{\normalfont{\textbf{not}} \normalfont{\text{success}}}{
                    $\mathcal{T}_{\text{\normalfont{curr}}} \gets \normalfont{\text{Explore Mode}}$\;\label{end inspect mode}
                }       
                {
                    \normalfont{\textbf{executeInspectionPath}}()\;
                    $\mathcal{T}_{\text{\normalfont{curr}}} \gets \normalfont{\text{Return Mode}}$\;
                }            }
            \Case{\normalfont{Return Mode}}{\label{start return mode}
                $\mathcal{P}_{\text{\normalfont{goal}}} \gets$ \normalfont{\textbf{getReturnGoal}}()\;
                $\sigma_{\normalfont{\text{traj}}} \gets$ \textbf{lowerLevelPlanner}($\mathcal{P}_{\text{\normalfont{goal}}}$)\;   \label{end inspection}
           }
    }

}

\end{algorithm}

\subsection{Perception and 3D Dynamic Mapping} \label{perception mapping}
\begin{figure}[t] 
    \centering
    \includegraphics[scale=0.575]{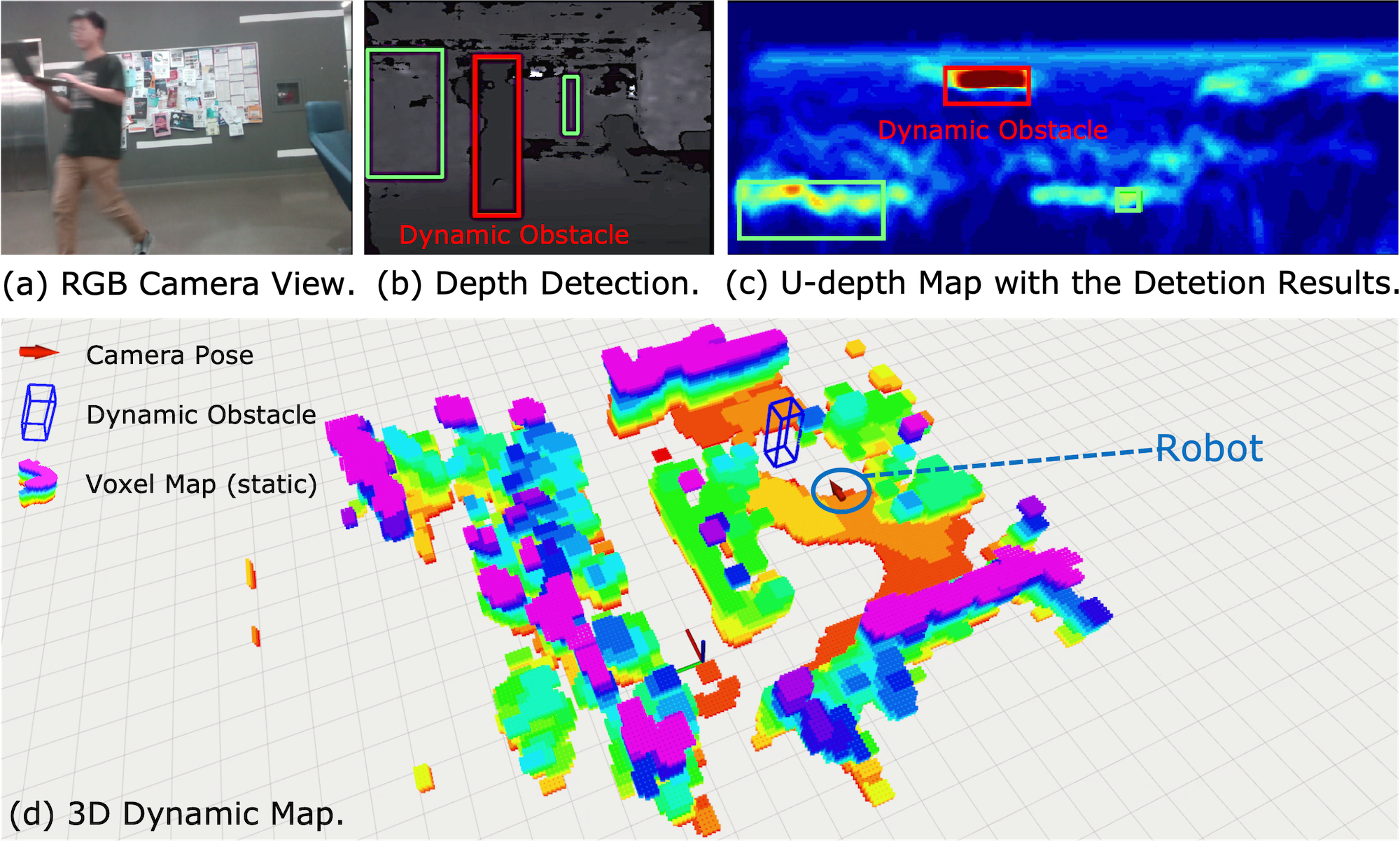}
    \caption{Illustration of the proposed 3D dynamic map. (a) A person walks in front of the robot in the RGB camera view. (b) The person is detected as a dynamic obstacle in the depth image. (c) The detection results in the U-depth map for obstacle widths and thicknesses. (d) The 3D dynamic map shows the dynamic obstacle as a bounding box and static obstacles as the voxel map.}
    \label{dynamic_map}
\end{figure}

This section introduces our proposed 3D dynamic map for navigating dynamic environments, as shown in Fig. \ref{dynamic_map}d. Our dynamic map adopts a hybrid method to represent environments by using the occupancy voxels for static obstacles and the bounding boxes for dynamic obstacles. For static obstacles, we predefine a static voxel map size (i.e., maximum voxel numbers) based on the environment and store the occupancy information of each voxel in an array with the preserved length. This allows our planners to access the occupancy information with $\mathcal{O}(1)$ time complexity. For the occupancy information update of each voxel, as most static occupancy mapping algorithm does, we apply the classic Bayesian filter with the Markov assumption:
\begin{equation}
l_{t}(x) = \log\frac{p(x|z_{t})}{p(\Bar{x}|z_{t})} + \log\frac{p(\Bar{x})}{p(x)} + l_{t-1}(x), \label{occupancy update}
\end{equation}
where $l_{t}(x)$ is the log odds for the voxel being occupied. By applying Eqn. \ref{occupancy update}, we can update the occupancy information (i.e., log odds) for each voxel by  recursively adding the inverse sensor model $\log\frac{p(x|z_{t})}{p(\Bar{x}|z_{t})}$ with the predefined prior $\log\frac{p(\Bar{x})}{p(x)}$. 
Besides, since dynamic obstacles can also be mapped into the static voxel map, which can lead to noisy voxels, we iterate through each detected dynamic obstacle bounding box and set all voxels inside the dynamic regions to be free.

The dynamic obstacles are detected and tracked using the depth image and represented by axis-aligned 3D bounding boxes. There are mainly three steps in the proposed method: region proposal detection, map-depth fusion and dynamic obstacle filtering. In the region proposal detection step, we use the method mentioned in \cite{reactiveUV} to generate the U-depth map, as shown in Fig. \ref{dynamic_map}c, by constructing a histogram of the depth values using the depth image. The vertical axis from top to bottom of the U-depth map represents the depth range of the user-defined bin width. Intuitively, the U-depth map can be viewed as a top-down view image. Inspired by \cite{reactiveUV}\cite{20_chance}, we apply the line grouping method to detect the obstacle regions in the U-depth map. With these detection results, we can obtain the widths and thicknesses of obstacles and then further find the corresponding heights in the original depth image as shown in Fig. \ref{dynamic_map}b. After this step, we can get the  \say{region proposal bounding boxes} for dynamic obstacles by applying coordinate transformation into the map frame. Since the region proposals are only the rough detection results, our second step, map-depth fusion, inflates those region proposals locally with a ratio $\lambda$ and then searches occupied voxels from the static voxel map to get the refined bounding boxes of obstacles. With the refined bounding boxes, the dynamic obstacle filtering method is applied to identify and track dynamic obstacles. First, we utilize the Kalman filter to track and compute the velocity of each obstacle bounding box with the linear propagation model: 
\begin{equation}
    \textbf{p}^{k+1}_{o} = \textbf{p}^{k}_{o} + \textbf{v}^{k}_{o}(t_{k+1} - t_{k}), \ \ \textbf{v}^{k}_{o} = \frac{\textbf{p}^{k}_{o} - \textbf{p}^{k-1}_{o}}{t_{k} - t_{k-1}},
\end{equation}
where $\textbf{p}^{k+1}_{o}$ is the predicted obstacle position in the next time step and $\textbf{v}^{k}_{o}$ is the previously estimated velocity. Then, we identify those bounding boxes with velocities greater than the threshold $V_{\text{th}}$ as the dynamic obstacles. Finally, we remove the bounding boxes with jerky motions using the obstacles' historical velocities, considering the detection noises that make static obstacles shake back and forth slightly. Because we use velocity as the criterion to identify dynamic obstacles, if a static obstacle begins to move, it will be detected as a dynamic obstacle represented by a bounding box. Consequently, the occupied voxels in the map are freed, and vice versa.

\subsection{Navigation and Obstacle Avoidance} \label{obstacle avoidance}
When a goal position is determined by the high-level task planner, the mid-level static planner first finds a smooth trajectory considering static obstacles. Then, using this static trajectory, the low-level dynamic planner optimizes a collision-free trajectory based on static and dynamic obstacles at a certain replanning frequency.  For the mid-level static planner, we apply the RRT* planner to find the waypoint path and use the minimum snap-based polynomial optimization with corridor constraints \cite{min_snap}\cite{poly} for trajectory generation. To achieve fast replanning for dynamic obstacle avoidance, the low-level planner adopts our gradient-based trajectory optimization. The B-spline trajectory with order k over a time knot vector can be parameterized as a series of control points:
\begin{equation}
\mathcal{\hat{S}} = \{\textbf{P}_{1}, \textbf{P}_{2}, \textbf{P}_{3}, ..., \textbf{P}_{N-1}, \textbf{P}_{N}\},  \  \   \textbf{P}_{i} \in \mathcal{R}^\text{3},
\end{equation}
where the optimization variable set $\mathcal{S}$ contains the $N-2(k-1)$ intermediate control points $\textbf{P}_{i}$. With the trajectory optimization variables, we can write the objective function as follows:
\begin{equation} \label{objective_function}
\begin{split}
    \text{C}_{\text{total}}(\mathcal{S}) = \alpha_{\text{control}} \cdot \text{C}_{\text{control}} + \alpha_{\text{smooth}} \cdot \text{C}_{\text{smooth}} \\ + \alpha_{\text{static}} \cdot \text{C}_{\text{static}} + \alpha_{\text{dynamic}} \cdot \text{C}_{\text{dynamic}},  
\end{split}
\end{equation}
and the weighted sum has four costs to minimize: the control limit cost, the smoothness cost, the static collision cost, and the dynamic collision cost. The control limit cost ensures the trajectory has feasible velocities and accelerations. The control points for velocity $\textbf{V}_{i}$ and acceleration $\textbf{A}_{i}$ are computed by:
\begin{equation}
    \textbf{V}_{i} = \frac{\textbf{P}_{i+1} - \textbf{P}_{i}}{\delta t}, \ \textbf{A}_{i} = \frac{\textbf{V}_{i+1} - \textbf{V}_{i}}{\delta t}, 
\end{equation}
where $\delta t$ is the time step. We use the L2 norm to penalize the infeasible velocities and accelerations:
\begin{equation}
    \text{C}_{\text{control}} = \sum_{i} {\frac{||\textbf{V}_{i} - \textbf{v}_{\text{max}}||_{2}^{2}}{\lambda_{\text{vel}}} + \frac{||\textbf{A}_{i} - \textbf{a}_{\text{max}}||_{2}^{2}}{\lambda_{\text{acc}}}}, \label{control_cost}
\end{equation}
in which $\textbf{v}_{\text{max}}$ and $\textbf{a}_{\text{max}}$ are the maximum velocity and acceleration limits. The $\lambda$ terms are the unit normalization factor. Note that the control limit costs are zero for velocities and acceleration that are less than the limits. The smoothness cost tries to reduce the jerk (i.e., the third derivative to position) of the trajectory using the following equations:
\begin{equation}
    \text{C}_{\text{smooth}} = \sum_{i} {||\textbf{J}_{i}||_{2}^{2}}, \ \  \textbf{J}_{i} = \frac{\textbf{A}_{i+1} - \textbf{A}_{i}}{\delta t}.
\end{equation}

\begin{figure}[t] 
    \centering
    \includegraphics[scale=0.85]{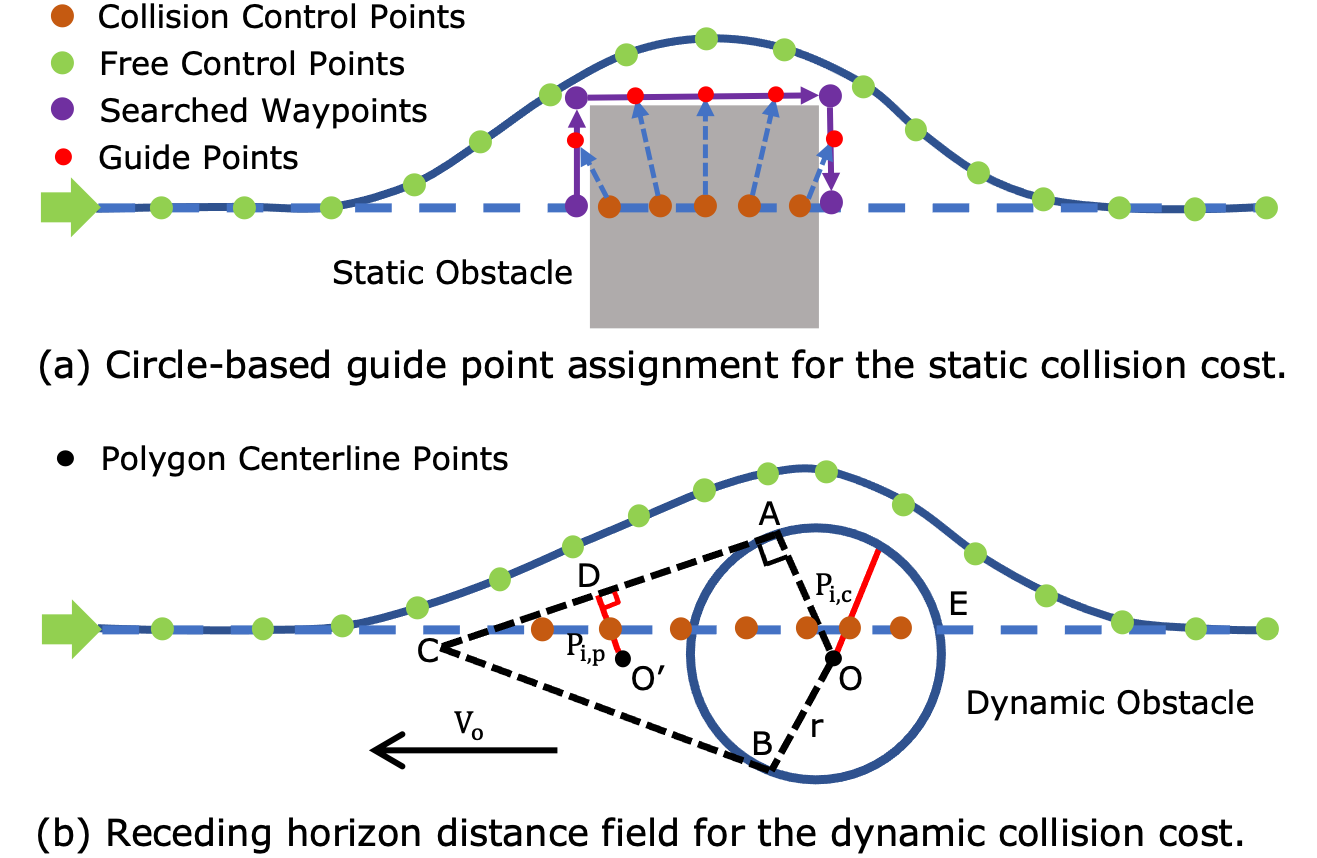}
    \caption{Illustration of the collision cost in our B-spline optimization. (a) The static collision cost is calculated using the proposed circle-based guide points (red dots). (b) The dynamic collision cost is obtained by the receding horizon distance field, which considers the future predictions of the obstacle positions.}
    \label{collision cost}
\end{figure}

The static collision cost is computed based on the proposed circle-based guide-point method shown in Fig. \ref{collision cost}a. The initial trajectory is shown as the blue dot line with the brown collision control points. To calculate the costs for those collision control points, we first search a collision-free path (purple dots and lines in Fig. \ref{collision cost}a) using A* or Dijkstra to bypass the static obstacle. If there are $N$ collision control points, we cast a ray for the collision control point of sequence order $n$ with the angle $\frac{180}{n+1}$ degree. Note that the angle is between the casting ray (dot blue arrow) and the line connecting the first and last collision control points. The guide points $\textbf{P}_{\text{guide}}$ are the intersection points of the casting ray with the searched path. The algorithm is circle-based because
the direction angles sweep a semi-circle. With the associated guide points for each collision control point, we design the total static collision cost based on experiments as a clipped cubic penalty function:
\begin{equation}
    \text{C}_{\text{static}} = \sum_{i} \biggl(\textbf{max}\Bigl(\text{d}_{\text{safe}} - \textbf{signDist}(\textbf{P}_{\text{i}}, \textbf{P}^{\text{i}}_{\text{guide}}), 0\Bigl)\biggl)^3, \label{static_cost_function}
\end{equation}
where $\text{d}_{\text{safe}}$ is the user-defined safe distance, and the signed distance function defines the positive and negative distance as the control point outside and inside the obstacle. Intuitively, we penalize the control points with small or negative distances to obstacles, and the static collision costs are zero for control points with a distance greater than the safe distance.   

Since the dynamic obstacles are moving, it is unreliable to only use the current detected information for cost computation. So, we propose the receding horizon distance field to estimate the dynamic collision cost with future predictions shown in Fig. \ref{collision cost}b. In this figure, the dynamic obstacle with left moving velocity $\text{V}_\text{o}$ is represented as the blue circle with the center $\text{O}$ and the radius $\text{r}$. We apply linear prediction to get the obstacle's future position $\text{C}$ with the prediction horizon $k$ time step. Since the reliability of future prediction decreases with the increasing prediction time, we linearly decrease the obstacle size to zero at the final predicted position $\text{C}$ in the receding horizon manner. So, we can obtain the collision region as the combination of a polygon region $\text{AOBC}$ and a circular region enclosed by the arc $\arc{\text{AEB}}$, line $\text{AO}$, and line $\text{BO}$. When the control point $\textbf{P}_{\text{i,p}}$ is inside the polygon region, we draw a red line through the control point $\textbf{P}_{\text{i,p}}$ perpendicular to the line $\text{AC}$ intersecting at point $D$. The distance $\text{d}_{\text{i}}$ to the safe area (outside the collision region) can be computed as:
\begin{equation}
    \Delta \text{d}_{\text{i}} = ||\textbf{D} - \textbf{O}^{'}||_{2} - ||\textbf{P}_{\text{i,p}} - \textbf{O}^{'}||_{2}. \label{case2}
\end{equation}
On the other hand, when the control point $\textbf{P}_\text{i,c}$ is inside the circular region, the  distance $\text{d}_{\text{i}}$ to the safe area is: 
\begin{equation}
    \Delta \text{d}_{\text{i}} = \text{r} - ||\textbf{P}_\text{i,c} - \textbf{O}_{0}||_{2}. \label{case1}
\end{equation}
For the control points $\textbf{P}_\text{i,out}$ that are outside both polygon and circular regions, we set the distance $\text{d}_{\text{i}}$ to the safe area to zero. So, with the distance to the safe area, we can use the following equation to compute the final dynamic collision cost:
\begin{equation}
    \text{C}_{\text{dynamic}} = \sum_{i} \Bigl(\textbf{max}(\Delta \text{d}_{\text{i}}, 0)\Bigl)^3. \label{dynamic_cost_function}
\end{equation}
For both static and dynamic collision costs, the gradients can be computed using the chain rule with Eqn. \ref{static_cost_function} and Eqn.\ref{dynamic_cost_function}.

\subsection{Inspection and 3D Reconstruction} \label{3d reconstruction}
After finishing the entire inspection task, the data post-processing module applies the Structure-from-Motion (SfM) to reconstruct the 3D shape of the inspection target from the collected target images. When the robot has reached the inspection target, it first explores the local area until having enough map information about the target. Then, the robot generates a zigzag pattern path with its camera facing toward the target wall and collects color images during the flight. After finishing the zigzag path, the robot will turn 45 degrees towards the corner of the target wall, following a rectangular shape path to collect images of the wall fringe. We adopted the zigzag pattern inspection path because it is a simple yet efficient way to cover the target surface, making it suitable for industrial applications. However, there are more intelligent ways to find the best viewpoints considering execution time and reconstruction quality, which are outside the scope of this paper. Our SfM pipeline for reconstruction is based on COLMAP \cite{colmap}. The algorithm first extracts the features of each image using a numerical descriptor. Since our input images are from the streaming of an RGB camera, the second step utilizes sequential matching to find the correspondence in different images. Finally, from an initial corresponding image pair, the algorithm incrementally reconstructs the 3D shape of the inspection target by triangulating new points.

\section{Result and Discussion}
\subsection{Implementation Details}
We conduct simulation experiments and physical flight tests in dynamic tunnel environments to evaluate the proposed method's performance. The simulation environments are based on ROS and Gazebo. For the physical experiments, we visited a tunnel under construction in Japan and applied our customized quadcopter (Fig. \ref{physical_flight} left) to test the proposed framework. The quadcopter is equipped with a RealSense D435i camera, a PX4-based flight controller, and an NVIDIA Xavier NX onboard computer. We adopt the visual-inertial odometry (VIO) algorithm for robot state estimation. All of the perception and planning computations are performed within the onboard computer. The color images are collected during the inspection stage with the RealSense D435i camera, and the data post-processing for 3D reconstruction is completed using the desktop with an NVIDIA RTX 3080 GPU.


\subsection{Evaluation of Navigation and Obstacle Avoidance}
\begin{figure}[t] 
    \centering
    \includegraphics[scale=0.405]{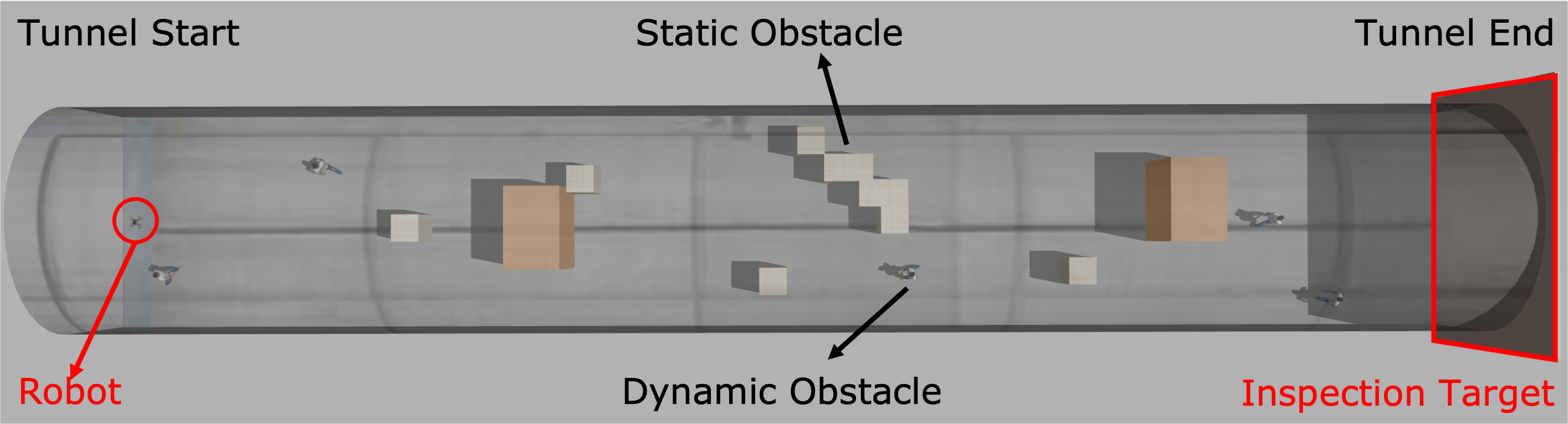}
    \caption{Illustration of an example simulation tunnel environment in Gazebo. In the forward task, the robot needs to navigate from the tunnel start (left side) to the tunnel end (right side) and avoid static and dynamic obstacles.}
    \label{simulation}
\end{figure}

The navigation and obstacle avoidance in the forward task (i.e., approaching the tunnel end) is the most challenging and time-consuming part of the entire inspection process since the environment is cluttered and unknown. So, to evaluate the performance of forward navigation and obstacle avoidance, we prepared 5 simulation environments containing different static and dynamic obstacles, with one example environment shown in Fig. \ref{simulation}. For benchmarking, we select the sampling-based planning methods (SBP) \cite{RH-NBV}\cite{tunnel_inspection} and the dynamic exploration planning (DEP) method \cite{DEP} with modifications to the tunnel environments. Besides, we also include our method without using the dynamic map (mentioned in Sec. \ref{perception mapping}) to compare the obstacle avoidance performance. In each experiment, we let the robot navigate from the start of the tunnel to the end of the tunnel. We run 10 experiments in each environment of different obstacles and record the average navigation time, the average replanning time for dynamic obstacle avoidance, and the collision rate over all experiments. Note that we set the navigation time and replanning time of the sampling-based planning methods (SBP) \cite{RH-NBV}\cite{tunnel_inspection} to $100\%$ for comparison. The collision rate is calculated by the number of experiments with collisions divided by the total number of experiments. 

From the results in Table \ref{benchmark}, one can see that our method has the second least navigation time, which is $81.69\%$ of the sampling-based planning (SBP) method, and takes almost the same amount of time as its non-dynamic-map version. The dynamic exploration planning (DEP) method uses less time than the sampling-based method and longer time than our method. From our observations, both the SBP and the DEP generate their trajectories inside the explored regions, which is over-conservative, leading to more stop-and-go behavior. On the contrary, since our planner adopts a hierarchical scheme, the task planner first tries using the more aggressive local planner for obstacle avoidance by planning in the unknown regions and only applies the conservative exploration planner when the local planning fails. This task-switching behavior hugely reduces the navigation time. For the replanning time, our method takes only $1.16\%$ of the time compared to the SBP and significantly less than the DEP. This huge difference in the replanning speed is mainly due to our computationally lightweight gradient-based trajectory optimization and the long computation time in the information gain evaluation of the SBP and the DEP. For the collision rate, it is shown that our method has no collision among all experiment runs, and both the SBP and our method without the dynamic map have a high collision rate (around $30\%$). The DEP has a lower collision rate than the SBP since it utilizes an incremental roadmap for faster dynamic obstacle avoidance but still has more collisions than our method. Comparing our method with and without the dynamic map shows that the dynamic map version has a much lower collision rate by using dynamic obstacle information.

\begin{table}[h]
\begin{center}
\caption{Comparison of Navigation and Obstacle Avoidance.} \label{benchmark}
\begin{tabular}{  l   c  c  c } 
 \hline

  Methods & Nav. Time & Replan. Time & Collision Rate  \Tstrut\\ 
 \hline
 SBP \cite{RH-NBV}\cite{tunnel_inspection}  & $100\pm0\%$  & $100\%$ & $30.00\%$ \Tstrut\\ 
 DEP \cite{DEP} & $92.80\pm3.01\%$ & $54.30\%$ & $24.00\%$ \\

 Ours w/o DM & $\textbf{81.06}\pm\textbf{4.40}\%$  & $1.20\%$ &  $32.00\%$\\

 \textbf{Ours} & $81.69\pm3.66\%$  & $\textbf{1.16}\%$ &  $\textbf{0.00}\%$\\  

 \hline
\end{tabular}
\end{center}
\end{table}

\subsection{Evaluation of Dynamic Obstacle Tracking}
We measure the average tracking errors in positions, velocities, and obstacle sizes shown in Table \ref{detection_result} to evaluate the dynamic obstacle detection and tracking performance. The ground truth states of the obstacles can be easily obtained in the simulation experiments, and we apply the OptiTrack motion capture system in the physical tests to obtain the ground truth states. We let two persons walk within the motion capture area, compare the tracking results from the robot and the motion capture system, and use the average value differences as tracking errors. From Table \ref{detection_result}, one can see that the position errors are $0.09$m and $0.19$m in simulation and physical tests, respectively. The position errors in the physical tests are larger than in simulation tests due to the image's noises from the depth camera. Similarly, the camera noises also make the velocity errors in physical tests greater than the simulations'. The size errors are similar in both simulation and physical tests. In the experiments, to account for the tracking errors in the positions, velocities, and sizes, we increase the safety distance to obstacles by a self-defined size $r$, and our experiment results prove that our dynamic obstacle tracking system can let successfully avoid moving obstacles.

\begin{table}[h]
\begin{center}
\caption{Measurement of the Detection and Tracking Errors.} \label{detection_result}

\begin{tabular}{|c | c | c|  } 
 \hline

  Errors & Simulation Tests & Physical Tests \Tstrut\\ 
 \hline

 Position Error (\text{m})  & $0.09$  & $0.19$ \Tstrut\\ 
 \hline

 Velocity Error (\text{m}/\text{s}) & $0.10$  & $0.21$ \Tstrut\\  
 \hline
 
 Size Error (\text{m}) & $0.25$  & $0.25$ \Tstrut\\  
 \hline
\end{tabular}
\end{center}
\end{table}

\subsection{Physical Flight Tests}
\begin{figure}[t] 
    \centering
    \includegraphics[scale=0.57]{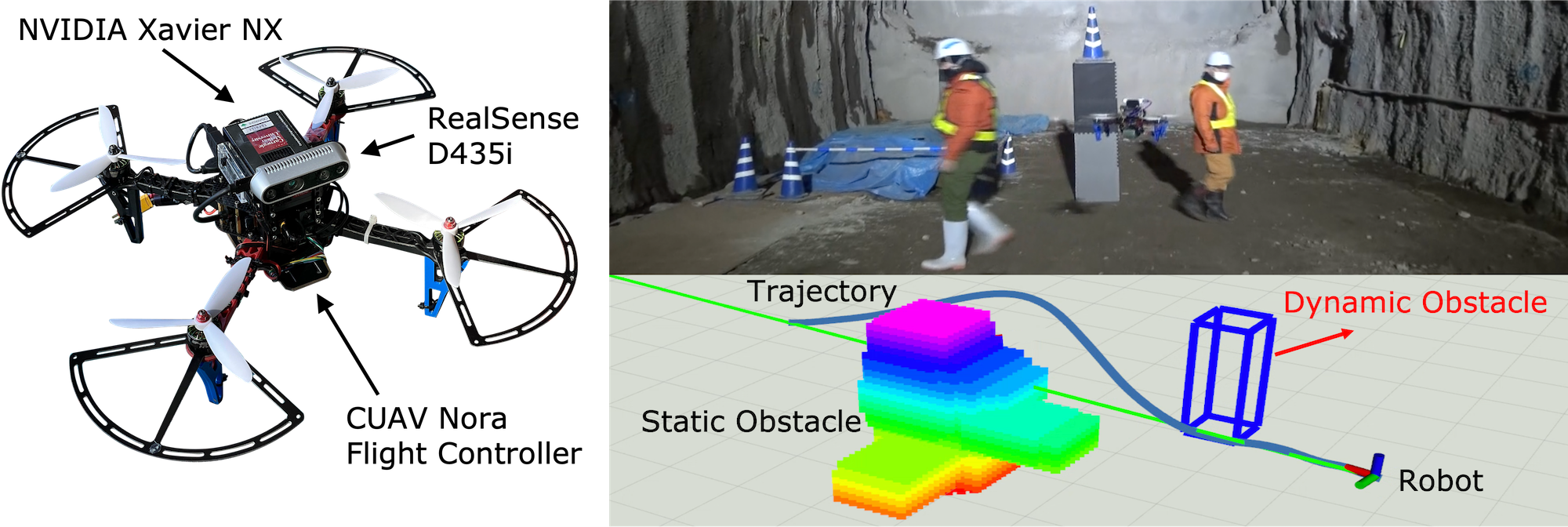}
    \caption{The inspection experiments with the customized quadcopter (left) in a tunnel under construction in Japan (top right). The bottom right visualizes the static and dynamic obstacles and the robot planned trajectory.}
    \label{physical_flight}
\end{figure}

To evaluate and verify the proposed framework, we ran flight tests in a tunnel under construction in Japan, shown in Fig. \ref{intro_figure} and \ref{physical_flight}. In each flight test, the robot starts at 20 meters in front of the tunnel excavation front and navigates toward the inspection area. Note that there are static and dynamic obstacles (i.e., walking workers) on the robot's way to its target location shown at the top of Fig. \ref{physical_flight}. The corresponding Rviz visualization is shown at the bottom of Fig. \ref{physical_flight}, and one can see that the robot can generate a collision-free trajectory for navigation. After reaching the inspection area, the robot will follow the zigzag path to inspect the tunnel excavation front shown in Fig. \ref{intro_figure}d and collect RGB images for further 3D reconstruction. During the navigation period, the robot's velocity is maintained at $1.0$m/s. The results show that our framework can complete the entire inspection task autonomously.


\subsection{Evaluation of 3D Reconstruction}
\begin{figure}[t] 
    \centering
    \includegraphics[scale=1.055]{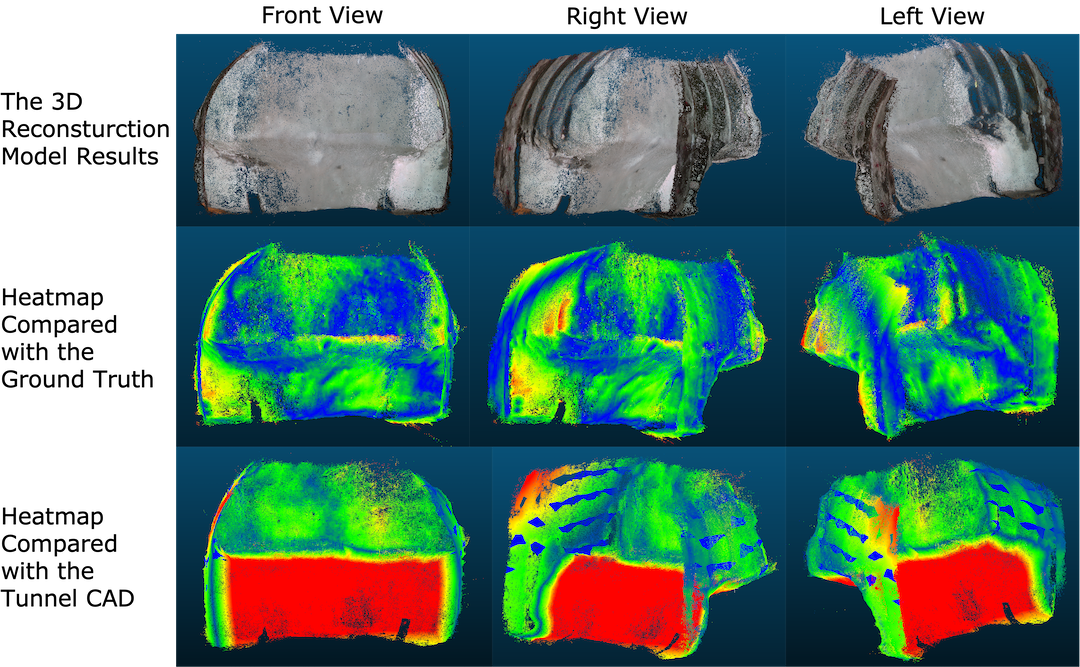}
    \caption{The 3D reconstruction results of the excavation front of the tunnel under construction in Japan. The first row shows the 3D reconstruction model from different views. The second row visualizes the error heatmap obtained from the comparison of the laser-scanned ground truth. The third row presents the heatmap comparison of the reconstruction model with the CAD model.}
    \label{reconstruction_result}
\end{figure}

The final output of our framework is the 3D shape of the tunnel excavation front shown in Fig. \ref{reconstruction_result}. To obtain the results, we run the SfM-based reconstruction mentioned in Sec. \ref{3d reconstruction} with 294 color images of 640x480 resolution. The total processing time is 30 minutes using an NVIDIA RTX 3080 GPU, and the minimum number of images required for this experiment is 60 images which take only 5 minutes for reconstruction.  In Fig. \ref{reconstruction_result}, the first row shows the reconstruction results from different views, and the second row visualizes the error heatmap from the comparison with the ground truth model. Note that we use the Topcon laser scanner to obtain the ground truth model of the inspection target. The red and blue portion of the heatmap represents the high and low reconstruction error values. The reconstruction model has an average error of 5.38cm, which is 0.67\% of the tunnel's width and height, with a standard deviation of 7.96cm. This reconstruction error falls within an acceptable range to perform construction operations based on our industrial partners' suggestions. The third row shows the heatmap comparison with the tunnel CAD model, the designed shape for the tunnel. From the heatmap, the workers can identify the yellow and red regions as the locations for concrete spraying and excavation.  

\section{Conclusion and Future Work}
This paper presents a vision-based autonomous UAV inspection framework for tunnel environments. The proposed framework adopts a hierarchical planning scheme to solve the complicated inspection problem using different planning layers. Our depth-based 3D dynamic map can represent static obstacles and track dynamic obstacles simultaneously. The experiment results prove that our framework can make the quadcopter safely navigate toward the inspection target to perform the inspection and return to the origin. The final 3D reconstruction results obtained from our SfM-based data post-processing pipeline have a low error compared to the ground truth. For future work, we want to apply learning-based methods to classify dynamic obstacles. Moreover, a more optimal inspection path than the current zigzag path could be explored using a lightweight online reconstruction method, which may lead to improved reconstruction quality.

\section{Acknowledgement}
\noindent The authors would like to thank TOPRISE CO., LTD and Obayashi Corporation for their financial support in this work and for providing a tunnel construction site for the flight tests. 

\bibliographystyle{IEEEtran}
\bibliography{IEEEabrv,bibliography.bib}

\begin{thebibliography}{10}
\providecommand{\url}[1]{#1}
\csname url@samestyle\endcsname
\providecommand{\newblock}{\relax}
\providecommand{\bibinfo}[2]{#2}
\providecommand{\BIBentrySTDinterwordspacing}{\spaceskip=0pt\relax}
\providecommand{\BIBentryALTinterwordstretchfactor}{4}
\providecommand{\BIBentryALTinterwordspacing}{\spaceskip=\fontdimen2\font plus
\BIBentryALTinterwordstretchfactor\fontdimen3\font minus
  \fontdimen4\font\relax}
\providecommand{\BIBforeignlanguage}[2]{{%
\expandafter\ifx\csname l@#1\endcsname\relax
\typeout{** WARNING: IEEEtran.bst: No hyphenation pattern has been}%
\typeout{** loaded for the language `#1'. Using the pattern for}%
\typeout{** the default language instead.}%
\else
\language=\csname l@#1\endcsname
\fi
#2}}
\providecommand{\BIBdecl}{\relax}
\BIBdecl

\bibitem{RH-NBV}
A.~{Bircher}, M.~{Kamel}, K.~{Alexis}, H.~{Oleynikova}, and R.~{Siegwart},
  ``Receding horizon "next-best-view" planner for 3d exploration,'' in
  \emph{2016 IEEE International Conference on Robotics and Automation (ICRA)},
  2016, pp. 1462--1468.

\bibitem{AEP}
M.~{Selin}, M.~{Tiger}, D.~{Duberg}, F.~{Heintz}, and P.~{Jensfelt},
  ``Efficient autonomous exploration planning of large-scale 3-d
  environments,'' \emph{IEEE Robotics and Automation Letters}, vol.~4, no.~2,
  pp. 1699--1706, 2019.

\bibitem{TSDF}
L.~{Schmid}, M.~{Pantic}, R.~{Khanna}, L.~{Ott}, R.~{Siegwart}, and J.~{Nieto},
  ``An efficient sampling-based method for online informative path planning in
  unknown environments,'' \emph{IEEE Robotics and Automation Letters}, vol.~5,
  no.~2, pp. 1500--1507, 2020.

\bibitem{DEP}
Z.~Xu, D.~Deng, and K.~Shimada, ``Autonomous uav exploration of dynamic
  environments via incremental sampling and probabilistic roadmap,'' \emph{IEEE
  Robotics and Automation Letters}, vol.~6, no.~2, pp. 2729--2736, 2021.

\bibitem{tunnel_inspection}
B.~Bendris and J.~Cayero~Becerra, ``Design and experimental evaluation of an
  aerial solution for visual inspection of tunnel-like infrastructures,''
  \emph{Remote Sensing}, vol.~14, no.~1, p. 195, 2022.

\bibitem{reactiveUV}
H.~Oleynikova, D.~Honegger, and M.~Pollefeys, ``Reactive avoidance using
  embedded stereo vision for mav flight,'' in \emph{2015 IEEE International
  Conference on Robotics and Automation (ICRA)}.\hskip 1em plus 0.5em minus
  0.4em\relax IEEE, 2015, pp. 50--56.

\bibitem{visionCCMPC}
J.~Lin, H.~Zhu, and J.~Alonso-Mora, ``Robust vision-based obstacle avoidance
  for micro aerial vehicles in dynamic environments,'' in \emph{2020 IEEE
  International Conference on Robotics and Automation (ICRA)}.\hskip 1em plus
  0.5em minus 0.4em\relax IEEE, 2020, pp. 2682--2688.

\bibitem{ViGO}
Z.~Xu, Y.~Xiu, X.~Zhan, B.~Chen, and K.~Shimada, ``Vision-aided uav navigation
  and dynamic obstacle avoidance using gradient-based b-spline trajectory
  optimization,'' \emph{arXiv preprint arXiv:2209.07003}, 2022.

\bibitem{bridge_GTSP}
P.~Shanthakumar, K.~Yu, M.~Singh, J.~Orevillo, E.~Bianchi, M.~Hebdon, and
  P.~Tokekar, ``View planning and navigation algorithms for autonomous bridge
  inspection with uavs,'' in \emph{International Symposium on Experimental
  Robotics}.\hskip 1em plus 0.5em minus 0.4em\relax Springer, 2020, pp.
  201--210.

\bibitem{model_based_1}
N.~Bolourian and A.~Hammad, ``Lidar-equipped uav path planning considering
  potential locations of defects for bridge inspection,'' \emph{Automation in
  Construction}, vol. 117, p. 103250, 2020.

\bibitem{model_based_2}
Y.~Tan, S.~Li, H.~Liu, P.~Chen, and Z.~Zhou, ``Automatic inspection data
  collection of building surface based on bim and uav,'' \emph{Automation in
  Construction}, vol. 131, p. 103881, 2021.

\bibitem{lidar_tunnel1}
T.~Elmokadem, ``A 3d reactive navigation method for uavs in unknown tunnel-like
  environments,'' in \emph{2020 Australian and New Zealand Control Conference
  (ANZCC)}.\hskip 1em plus 0.5em minus 0.4em\relax IEEE, 2020, pp. 119--124.

\bibitem{lidar_tunnel2}
T.~Elmokadem and A.~V. Savkin, ``A method for autonomous collision-free
  navigation of a quadrotor uav in unknown tunnel-like environments,''
  \emph{Robotica}, vol.~40, no.~4, pp. 835--861, 2022.

\bibitem{tunnel_reconstruction}
R.~S. Pahwa, K.~Y. Chan, J.~Bai, V.~B. Saputra, M.~N. Do, and S.~Foong, ``Dense
  3d reconstruction for visual tunnel inspection using unmanned aerial
  vehicle,'' in \emph{2019 IEEE/RSJ International Conference on Intelligent
  Robots and Systems (IROS)}.\hskip 1em plus 0.5em minus 0.4em\relax IEEE,
  2019, pp. 7025--7032.

\bibitem{nbv_idea}
C.~Connolly, ``The determination of next best views,'' in \emph{Proceedings.
  1985 IEEE international conference on robotics and automation}, vol.~2.\hskip
  1em plus 0.5em minus 0.4em\relax IEEE, 1985, pp. 432--435.

\bibitem{first_frontier}
B.~Yamauchi, ``A frontier-based approach for autonomous exploration,'' in
  \emph{Proceedings 1997 IEEE International Symposium on Computational
  Intelligence in Robotics and Automation CIRA'97.'Towards New Computational
  Principles for Robotics and Automation'}.\hskip 1em plus 0.5em minus
  0.4em\relax IEEE, 1997, pp. 146--151.

\bibitem{rapid_frontier}
T.~Cieslewski, E.~Kaufmann, and D.~Scaramuzza, ``Rapid exploration with
  multi-rotors: A frontier selection method for high speed flight,'' in
  \emph{2017 IEEE/RSJ International Conference on Intelligent Robots and
  Systems (IROS)}.\hskip 1em plus 0.5em minus 0.4em\relax IEEE, 2017, pp.
  2135--2142.

\bibitem{information_gain}
B.~Charrow, G.~Kahn, S.~Patil, S.~Liu, K.~Goldberg, P.~Abbeel, N.~Michael, and
  V.~Kumar, ``Information-theoretic planning with trajectory optimization for
  dense 3d mapping.'' in \emph{Robotics: Science and Systems}, vol.~11, 2015,
  pp. 3--12.

\bibitem{apf}
O.~Khatib, ``Real-time obstacle avoidance for manipulators and mobile robots,''
  in \emph{Autonomous robot vehicles}.\hskip 1em plus 0.5em minus 0.4em\relax
  Springer, 1986, pp. 396--404.

\bibitem{orca}
J.~v.~d. Berg, S.~J. Guy, M.~Lin, and D.~Manocha, ``Reciprocal n-body collision
  avoidance,'' in \emph{Robotics research}.\hskip 1em plus 0.5em minus
  0.4em\relax Springer, 2011, pp. 3--19.

\bibitem{VO}
P.~Fiorini and Z.~Shiller, ``Motion planning in dynamic environments using
  velocity obstacles,'' \emph{The international journal of robotics research},
  vol.~17, no.~7, pp. 760--772, 1998.

\bibitem{11_chance}
L.~Blackmore, M.~Ono, and B.~C. Williams, ``Chance-constrained optimal path
  planning with obstacles,'' \emph{IEEE Transactions on Robotics}, vol.~27,
  no.~6, pp. 1080--1094, 2011.

\bibitem{19_chance}
H.~Zhu and J.~Alonso-Mora, ``Chance-constrained collision avoidance for mavs in
  dynamic environments,'' \emph{IEEE Robotics and Automation Letters}, vol.~4,
  no.~2, pp. 776--783, 2019.

\bibitem{20_chance}
J.~Lin, H.~Zhu, and J.~Alonso-Mora, ``Robust vision-based obstacle avoidance
  for micro aerial vehicles in dynamic environments,'' in \emph{2020 IEEE
  International Conference on Robotics and Automation (ICRA)}.\hskip 1em plus
  0.5em minus 0.4em\relax IEEE, 2020, pp. 2682--2688.

\bibitem{dpmpc}
Z.~Xu, D.~Deng, Y.~Dong, and K.~Shimada, ``Dpmpc-planner: A real-time uav
  trajectory planning framework for complex static environments with dynamic
  obstacles,'' in \emph{2022 International Conference on Robotics and
  Automation (ICRA)}.\hskip 1em plus 0.5em minus 0.4em\relax IEEE, 2022, pp.
  250--256.

\bibitem{chen2022risk}
G.~Chen, P.~Peng, P.~Zhang, and W.~Dong, ``Risk-aware trajectory sampling for
  quadrotor obstacle avoidance in dynamic environments,'' \emph{arXiv preprint
  arXiv:2201.06645}, 2022.

\bibitem{min_snap}
D.~Mellinger and V.~Kumar, ``Minimum snap trajectory generation and control for
  quadrotors,'' in \emph{2011 IEEE international conference on robotics and
  automation}.\hskip 1em plus 0.5em minus 0.4em\relax IEEE, 2011, pp.
  2520--2525.

\bibitem{poly}
C.~Richter, A.~Bry, and N.~Roy, ``Polynomial trajectory planning for aggressive
  quadrotor flight in dense indoor environments,'' in \emph{Robotics
  research}.\hskip 1em plus 0.5em minus 0.4em\relax Springer, 2016, pp.
  649--666.

\bibitem{colmap}
J.~L. Sch\"{o}nberger and J.-M. Frahm, ``Structure-from-motion revisited,'' in
  \emph{Conference on Computer Vision and Pattern Recognition (CVPR)}, 2016.

\end{thebibliography}

\end{document}